\documentclass[runningheads]{llncs}

\usepackage{eccv}

\usepackage{eccvabbrv}

\usepackage{graphicx}
\usepackage{booktabs}
\usepackage{multirow}
\usepackage{textcomp}

\usepackage[dvipsnames]{xcolor}

\usepackage[table]{xcolor}

\usepackage[accsupp]{axessibility}  

\usepackage{hyperref}

\usepackage{orcidlink}

\begin{document}

\title{Towards Context-Aware Clinical Motion Understanding in Daily Living at Home: Freezing of Gait Detection with Egocentric Vision} 

\titlerunning{Freezing of Gait Detection with IMU and Egocentric Vision}

\author{Vayalet Stefanova\inst{1}\orcidlink{0009-0006-0137-5033}$^{\star}$ \and
Diwas Lamsal\inst{1}\orcidlink{0009-0006-1125-2299}$^{\star}$ \and
Margot Genbrugge\inst{2}\orcidlink{0009-0002-8380-7290} \and
Maxim Yudayev\inst{1}\orcidlink{0000-0001-9521-1537} \and
Christian Schlenstedt\inst{3}\orcidlink{0000-0002-3838-6848} \and 
Moran Gilat\inst{2}\orcidlink{0000-0002-0072-4637} \and
Bart Vanrumste\inst{1}\orcidlink{0000-0002-9409-935X}$^{\dagger}$ \and
Benjamin Filtjens\inst{4, 5}\orcidlink{0000-0003-2609-6883}$^{\dagger}$ }

\authorrunning{V.~Stefanova, D.~Lamsal et al.}

\institute{Department of Electrical Engineering (ESAT), KU Leuven, Leuven, Belgium \and
Neurorehabilitation Research Group (eNRGy), Department of Rehabilitation Sciences, KU Leuven, Leuven, Belgium \and 
Institute of Interdisciplinary Exercise Science and Sports Medicine, Medical School Hamburg, Hamburg, Germany \and
Department of Engineering Systems and Services, Delft University of Technology, Delft, The Netherlands \and
Institute for Health Systems Science, Delft University of Technology, Delft, The Netherlands \\
\email{vayalet.stefanova@kuleuven.be, diwas.lamsal@kuleuven.be}}

\maketitle
\def\thefootnote{}
\footnotetext{$^{\star}$Equal contribution \quad $^{\dagger}$Equal senior contribution}
\def\thefootnote{\arabic{footnote}}

\begin{abstract}
Understanding motion in daily living requires context beyond kinematics, because similar inertial patterns during activities of daily living (ADLs) can reflect intentional stopping, object interaction, or pathological movement impairment. Egocentric vision provides task-related context that may help disambiguate these cases. We investigate this challenge through freezing of gait (FOG) detection in Parkinson’s disease (PD), a symptom strongly influenced by contextual factors during ADLs. Using synchronized egocentric video, wearable IMUs, and expert-annotated FOG labels collected from 13 PD participants in their homes, we evaluate frozen representations from pretrained ego-video and time-series foundation models, alongside an IMU-based TCN trained from scratch, under leave-one-subject-out evaluation. The IMU-based TCN achieved the strongest event-detection performance, reaching 42.3 F1 and 83.0 AUROC, compared with 32.6 F1 and 77.2 AUROC for V-JEPA2 ego-video features. Although ego-video alone did not outperform IMU-based sensing, it showed above-chance discrimination, and qualitative analyses suggest that egocentric vision may capture FOG-relevant information independent of IMUs. Together, these results support the use of pretrained ego-video representations to add contextual information to wearable-sensor-based clinical motion understanding in daily living.
\end{abstract}

\section{Introduction}
\label{sec:intro}

Understanding clinical motion in daily living environments is an important challenge for wearable and computer vision systems \cite{filtjens2026perspectives}. Unlike structured laboratory assessments, daily-life activities performed at home are highly variable and shaped by the surrounding environment and ongoing tasks. Commonly used kinematic sensors may be insufficient for determining clinical motion, as similar low-motion inertial patterns can arise from intentional stopping, object interaction, or balance recovery \cite{filtjens2026perspectives, yang2025multimodal, moore2023enhancing}. This ambiguity is most consequential for clinical applications that rely on accurate motion assessment and real-time closed-loop interventions for gait disorders \cite{kluge2021consensus, ginis2018cueing, melbournekinematic, van2020exoskeleton, moore2023enhancing}. Examples include continuous estimation of digital mobility outcomes \cite{kluge2021consensus}, exoskeletons, on-demand cueing, kinematic-based deep brain stimulation  \cite{ginis2018cueing, melbournekinematic, van2020exoskeleton}, and fall-risk assessment \cite{moore2023enhancing}.

Egocentric vision (ego-vision) is an emerging modality for capturing contextual motion information during activities of daily living (ADLs) \cite{egoexo4d}. By recording the wearer's first-person perspective, it directly observes the surrounding environment and ongoing activities. Ego-vision has been successfully applied to tasks such as action recognition and video captioning in home environments and is supported by the availability of large-scale benchmark datasets \cite{egoexo4d, damen2022rescaling}. The modality has also shown value in clinical applications such as fall-risk assessment in older adults and people with PD \cite{nouredanesh2022egocentric, moore2023enhancing, moore2024using}. The findings suggest that contextual information may be particularly relevant for movement disorders that strongly depend on environmental factors. 

An example of such disorder is freezing of gait (FOG) in Parkinson's disease (PD), defined as ``paroxysmal episodes wherein there is an inability to step effectively, despite attempting to do so'' \cite{gilat2026updated}. FOG is a major cause of falls in this population \cite{falls_bloem}. Although an episodic phenomenon, context during ADLs is particularly important for FOG, as freezing episodes are often triggered by environmental and cognitive factors. Examples that evoke FOG include turning, narrow passages (\eg, going through a doorway), cluttered pathways, and dual-tasking (\eg, carrying a full cup of water while walking) \cite{conde2023triggers}. Many patients also report recurrent FOG at specific locations (\ie, hotspots) within their homes \cite{zoetewei2024gait}. Despite the importance of context, automatic FOG detection has primarily relied on physiological and kinematic sensing modalities, including electroencephalography (EEG), electromyography (EMG), skin conductance (SC), heart rate signals, and the most commonly used inertial measurement units (IMUs) \cite{bansal2023techniques, mancini2025technology}. A key limitation of IMU-based FOG detection is the difficulty of distinguishing FOG episodes from voluntary stopping \cite{yang2025multimodal}. Whether ego-vision can address this challenge and support clinical motion understanding for FOG detection remains unexplored.

Another challenge specific to the clinical domain is the scarcity of large labeled home-based datasets, which makes training robust deep learning (DL) models from scratch difficult \cite{mancini2019clinical}. Foundation models (FMs) offer a promising alternative by using representations learned from large-scale video and time-series data, potentially reducing the need for task-specific clinical training data~\cite{liang2024foundation}. Frozen FM representations also enable lightweight comparisons across modalities by only training a simple classifier (\eg, linear probe). This allows to evaluate whether general-purpose features encode clinically relevant information while avoiding computationally expensive end-to-end retraining \cite{bommasani2021opportunities, adeli2025carepd}. Large-scale egocentric video datasets have further enabled FMs specialized in ego-vision representation learning \cite{pei2024egovideo,vinod2025egovlm}. In the context of FOG, this raises the question of whether pretrained ego-vision representations capture predictive features relevant to freezing.

Our contribution is a problem formulation and empirical study of context-aware clinical motion understanding in daily living at home. First, we introduce egocentric video as a contextual modality for FOG detection during home-based ADL. Second, we evaluate frozen video and IMU FM representations under subject-independent leave-one-subject-out (LOSO) evaluation on synchronized ego-video, IMU, and expert-annotated FOG data. Third, we show that temporal egocentric video representations contain predictive information for FOG, though less strong than IMU representations alone. Finally, through a qualitative analysis of modality-specific errors, we identify cases where visual context could help and where it fails.

\section{Related Work}

\subsection{Automated FOG detection}
 Expert annotations of video recordings serve as the gold-standard ground truth for identifying FOG episodes \cite{gilat2026updated}. Various machine learning and DL techniques have been proposed to learn representations across modalities, including end-to-end architectures based on Long Short-Term Memory (LSTM) \cite{mir2024parkinson}, convolutional \cite{yang2024automatic} and transformer networks \cite{koltermann2024gait}, as well as conventional classifiers such as SVMs and XGBoost \cite{rodriguez2017home, shi2022detection}. Among these, temporal convolutional networks (TCNs) \cite{lea2017temporal} are a popular choice for modeling inertial time-series data. Because of their ease of use most state-of-the-art FOG detection systems rely exclusively on IMU signals commonly acquired from the lower limbs, waist, or trunk \cite{mancini2019clinical}. Pose-based approaches, such as those using graph-based architectures \cite{filtjens2022automated}, instead exploit skeletal structure from marker-based or third-person motion capture to encode joint-level spatial relations beyond what raw IMU signals provide. However, these require an external body-view camera unavailable in egocentric setups such as ours, placing them outside the scope of this paper. Combining multiple sensing modalities has also been studied with the goal to improve detection robustness \cite{mesin2022multi, yang2025multimodal, zhang2022multimodal, wang2020freezing}. While these approaches provide promising results, their generalizability beyond standardized laboratory settings remains unclear. Additionally, to the best of our knowledge, no existing study has explored egocentric vision as a modality for FOG detection.
 
\subsection{Contextual understanding through ego-vision}
In recent years, ego-vision has been increasingly adopted to characterize the environmental and behavioural context of daily activities. Previous studies have used ego-vision to estimate fall risk in older adults by identifying walking surfaces and environmental hazards during daily life \cite{nouredanesh2022egocentric}. More recently, egocentric videos have been used for assessing fall risk in people with PD to augment IMU-based mobility measures with environmental context \cite{moore2023enhancing, moore2024using}. The authors fine-tuned a YOLOv8 \cite{yolov8} object detection model to identify the overlap of objects and the direct walking path of the subject to inform fall risk. Within the FOG domain, a recent study explored detection of environmental triggers commonly associated with FOG onset, such as tight turns, narrow passages, and floor pattern changes \cite{qian2025trigger}. The authors finetuned multimodal FMs on egocentric videos and textual descriptions of these trigger scenarios to enable a wearable trigger detection system. This addresses a related but distinct problem from ours: identifying contextual situations that may provoke FOG, rather than detecting the occurrence of a FOG episode itself. Moreover, validation in people with PD was not performed \cite{qian2025trigger}.

\subsection{Foundation models for video and wearable sensing}
FMs have become a common approach for representation learning in both video and time-series analysis. In the visual domain, models such as VideoMAE \cite{tong2022videomae}, V-JEPA~2 \cite{assran2025vjepa2}, and VideoPrism \cite{zhao2024videoprism} are trained on large-scale collections of third-person videos using self-supervised objectives. These models have shown strong transferability across a wide range of downstream tasks, including human activity recognition, captioning, and question answering \cite{zhang2024vision}. However, third-person and egocentric videos differ in viewpoint, motion patterns, and the visual context available to the camera. Consequently, recent works such as EgoVLM \cite{vinod2025egovlm} and EgoVideo \cite{pei2024egovideo} have been pretrained on large-scale egocentric datasets to better capture first-person representations. 

Similar advances have been made for processing time-series data. FMs such as UniMTS \cite{zhang2024unimts} and Chronos-2 \cite{ansari2025chronos2} are pretrained on large and diverse collections of temporal signals and have achieved strong performance across tasks including classification, forecasting, and anomaly detection. Importantly, such models have shown the ability to learn transferable representations from wearable sensing modalities such as accelerometer and gyroscope signals, making them a useful tool for IMU-based movement analysis \cite{zhang2024unimts, xu2021limu}. 

Building on these advances, we evaluate a representative set of FMs covering complementary design choices. For IMU signals, we use UniMTS \cite{zhang2024unimts}, a motion-specialized model pretrained for human activity recognition, and Chronos-2 \cite{ansari2025chronos2}, a general-purpose multivariate time-series model. For ego-video, we select models differing in temporal coverage, pretraining domain, and capacity: the single-frame image model DINOv3 \cite{dinov3}; the third-person video models VideoMAE-v2 \cite{wang2023videomaev2} and V-JEPA~2 \cite{assran2025vjepa2}, which encode short and long temporal windows, respectively; and the egocentric video model EgoVideo \cite{pei2024egovideo}.

\section{Methods}

\subsection{Dataset}

This study recruited 15 individuals with PD (5 female, 10 male) who reported experiencing FOG on a daily basis. Participants provided written informed consent before any study related activities, and the study was approved by the ethical committee of KU Leuven and the university hospital UZ Leuven (EC Research UZ/KU Leuven) with reference number S70220. The cohort had a mean age of $68.2$ years and mean disease duration of $13.5$ years. Demographic and clinical characteristics are summarized in \Cref{tab:participants}. Two participants were excluded from the analysis due to incomplete recordings.

\begin{table}[tb]
\caption{Participant demographics, clinical characteristics, and dataset statistics ($n=15$). Values are reported as mean $\pm$ standard deviation. The reported time spent frozen (\%TF) and number of FOG episodes (\#FOG) are calculated across both ON and OFF medication states.}
\label{tab:participants}
  \centering
  \renewcommand{\arraystretch}{1.2}
  \setlength{\tabcolsep}{4pt}
  \begin{tabular}{@{}|l|c|@{}}
    \hline
    \textbf{Characteristic} & \textbf{Mean} $\pm$ \textbf{SD} \\
    \hline
    Age (years) & 68.2 $\pm$ 8.8 \\
    Disease duration (years) & 13.5 $\pm$ 7.0 \\
    MDS-UPDRS III (ON) & 32.3 $\pm$ 10.9 \\
    MDS-UPDRS III (OFF) & 39.2 $\pm$ 11.0 \\
    MoCA & 25.8 $\pm$ 2.3 \\
    NFOGQ & 18.7 $\pm$ 4.2 \\
    \hline
    Recording duration (min) & 13.19 $\pm$ 5.63 \\
    \%TF (ADL) & 7.2 $\pm$ 6.5 \% \\
    \#FOG (ADL) & 14.4 $\pm$ 18.0 \\
    \hline
  \end{tabular}
\end{table}

\begin{figure*}[tb]
  \centering
  \begin{subfigure}{0.45\textwidth}
    \centering
    \includegraphics[width=\linewidth]{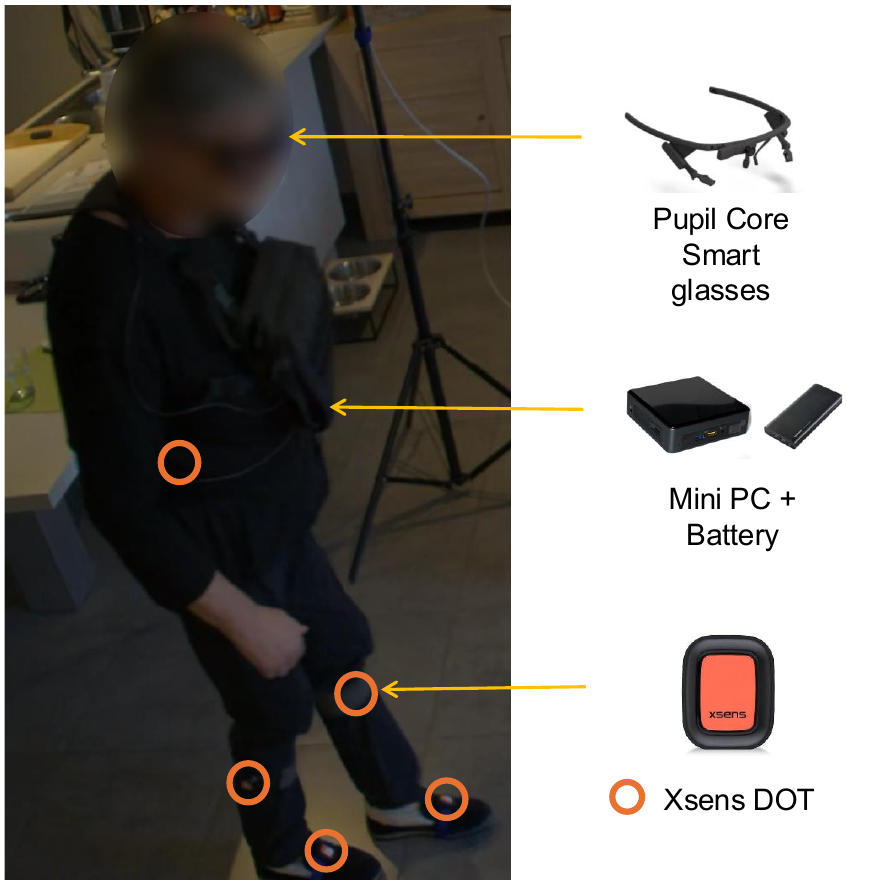}
    \caption{Sensor setup.}
    \label{fig:subject_setup}
  \end{subfigure}
  \hfill
  \begin{subfigure}{0.54\textwidth}
    \centering
    \includegraphics[width=\linewidth]{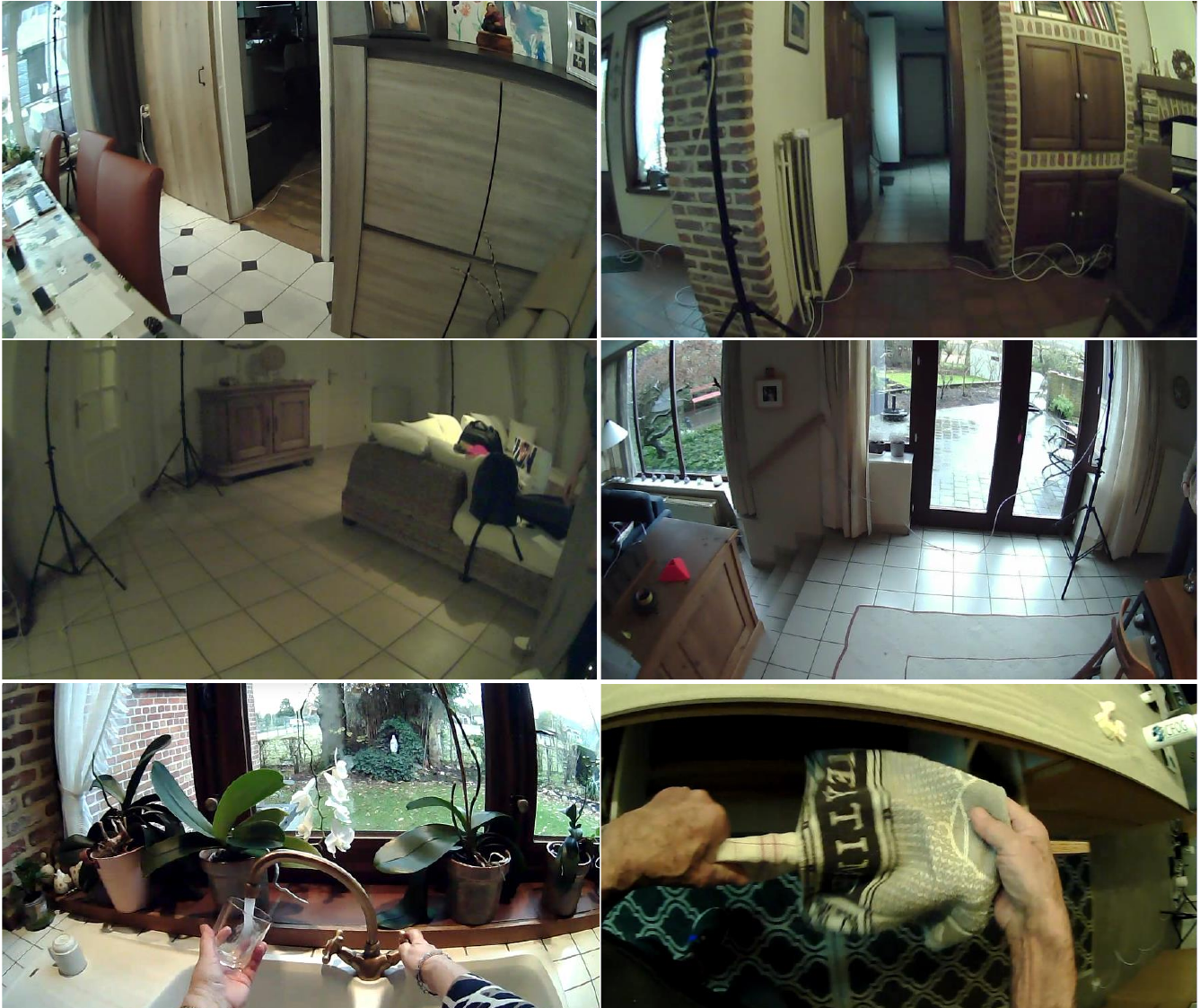}
    \caption{Example egocentric views.}
    \label{fig:ego_views}
  \end{subfigure}
  \caption{(a) Sensor setup for data collection. Pupil Core smart glasses for egocentric video and five Xsens DOT IMUs for lower-body inertial data, collected via a battery-powered mini PC. (b) Representative egocentric views captured during the ADL tasks: doorway (top), hotspot (middle), daily life (bottom).}
  \label{fig:data_collection_setup}
\end{figure*}

Subjects completed two data collection sessions in their homes, one in the OFF-medication state (following overnight withdrawal) and one ON-medication (1 hour after medication intake). Each session followed the same protocol, comprising five ADL tasks: walking through a doorway, two daily-life tasks (\eg, watering plants, washing dishes) randomly selected from a predefined list of 14 tasks (see Supplementary Material for the full list), and two hotspot tasks corresponding to locations within the home where the participant reported frequently experiencing FOG. These tasks were selected to reflect real-world conditions. Representative egocentric views of the three ADL task types (doorway, hotspot, and daily life) are shown in \cref{fig:ego_views}.

Data were collected using five Xsens DOT IMUs (60~Hz) positioned on the pelvis, bilateral shins, and the dorsum of both feet~(\cref{fig:subject_setup}). Ego-video was captured using Pupil Core smart glasses (30~Hz, $1280\times720$), while four external cameras (30~Hz, $2560\times1440$) were used for offline annotation. FOG episodes and tasks were post-hoc annotated by a human expert using the ELAN software \cite{gilat2019annotate}. Annotations followed the updated FOG definition, specifically the technical definition for scoring FOG from video recordings \cite{gilat2026updated}. Participants wore a lightweight chest pack containing an Intel NUC13ANKi7 mini-PC and a portable battery. We used HERMES \cite{HERMES} for data collection, which provided us with synchronized data across all modalities.

\begin{figure}[tb]
  \centering
  \includegraphics[height=5cm]{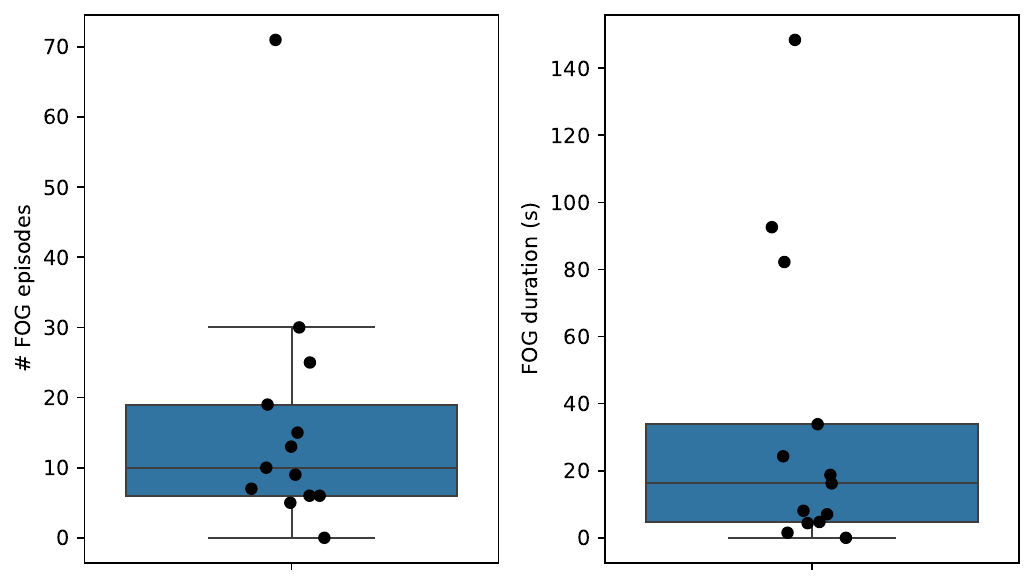}
  \caption{Inter-subject variability in FOG occurrence during home-based ADL. Each black point represents one participant, aggregated across the three ADL task types (doorway, hotspot, and daily life) and medication states (ON and OFF). Left: total number of FOG episodes per participant. Right: cumulative FOG duration per participant.}
  
  \label{fig:data}
\end{figure}

The final dataset for processing comprised 13 subjects and $171.6$ minutes of recordings. Data were segmented into individual tasks by removing transitions between activities. For each subject, ON and OFF medication sessions were combined to increase the number of windows and better represent real-life conditions, where FOG can occur during ADLs in both states. We observed considerable inter-subject variability in both the frequency and duration of FOG episodes (\cref{fig:data}). While some participants displayed little or no FOG, others experienced frequent and prolonged episodes. Such heterogeneity is in line with the clinical presentation of FOG, but also poses a challenge for training robust subject-independent FOG detection models \cite{yang2026deep}.

\subsection{Data preprocessing}
The data were segmented into overlapping windows using a fixed stride of 0.5\,s (see Supplementary Material for stride-sensitivity ablation). We use a 2\,s window as our main configuration, in line with common practices in window-based FOG detection \cite{SIGCHA2022105482}, and additionally evaluate 3\,s and 10\,s windows as ablations on a subset of models; the 10\,s length matches the temporal context used by the UniMTS foundation model during pretraining \cite{zhang2024unimts}.

Each window was assigned a binary FOG label and marked as FOG if it contained at least 0.5\,s of annotated FOG, a threshold chosen to preserve short FOG episodes, as the median episode duration in the dataset was 0.9\,s. For evaluation, each window was further assigned to whichever ADL task (doorway, hotspot, or daily life) represented more than 50\% of its duration. Windows with more than 50\% missing annotations, caused by occasional feet occlusion in the external camera views, were discarded.

\begin{figure}[tb]
    \centering
    \includegraphics[width=1\linewidth]{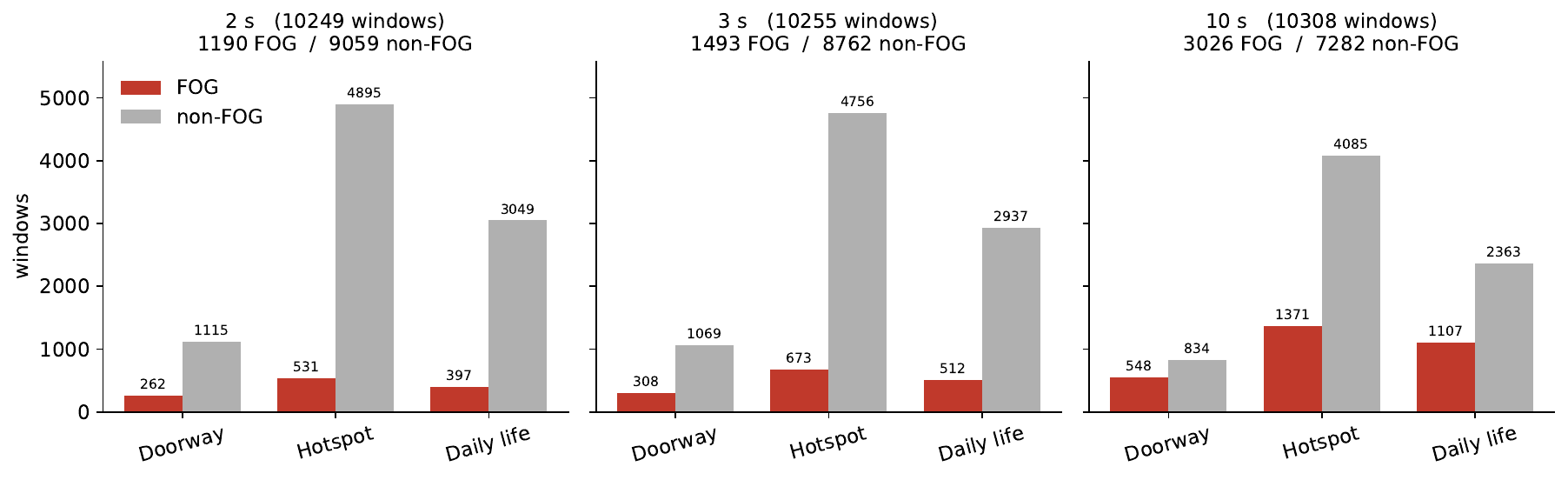}
    \caption{Window-level FOG vs.\ non-FOG counts per ADL task (doorway, hotspot, daily life) for the three window lengths, pooled over the 13 cohort subjects.}
    \label{fig:total_windows}
\end{figure}

\Cref{fig:total_windows} summarizes the resulting window counts per task and window length across the 13 subjects. Because the 0.5\,s FOG criterion is more easily satisfied, the number of FOG windows grows with window length, while the total number of retained windows varies because the task and missing-annotation criteria produce a different number of windows at each length. We therefore treat the three lengths as separate operating points rather than directly comparable conditions.

\subsection{Feature extraction}
For each window, modality-specific features were obtained from pretrained FMs, using UniMTS \cite{zhang2024unimts} and Chronos-2 \cite{ansari2025chronos2} for the IMU signals and VideoMAE-v2 \cite{wang2023videomaev2}, V-JEPA~2 \cite{assran2025vjepa2}, EgoVideo \cite{pei2024egovideo}, and DINOv3 \cite{dinov3} for the egocentric video.

\subsubsection{IMU features:}

First, missing IMU samples were linearly interpolated with edge-value extrapolation before resampling the signals to a uniform 60\,Hz grid. The resulting signals were then preprocessed according to the requirements of each FM. 

UniMTS is designed specifically for motion understanding and human activity recognition, using a spatio-temporal graph encoder that models body kinematics through a skeletal representation. Following the preprocessing recommended by the authors, the five IMU acceleration signals were mapped to the corresponding lower-body joints of the 22-joint SMPL skeletal representation \cite{ loper2023smpl} expected by the model and resampled from 60 \,Hz to 20 \,Hz using FFT-based resampling. The resulting signals were passed through the pretrained encoder and the output embedding was extracted for each window, yielding an embedding tensor of size $(N,D)$, where $N$ denotes the number of windows and $D=512$ is the feature embedding dimension. 

Chronos-2 was applied to both accelerometer and gyroscope signals. Each window was represented as a multivariate time series of the 30 IMU channels (5 IMUs $\times$ 6 channels) and passed through the pretrained model. This resulted in an embedding tensor $\mathbf{E}\in\mathbb{R}^{N\times C\times P\times D}$, where $N$ denotes the number of windows, $C=30$ the number of IMU channels, $P$ the number of temporal patches, and $D=768$ the latent embedding dimension. As Chronos generates a variable number of output tokens depending on the input sequence length, $P\in\{8,12,38\}$ for 2\,s, 3\,s, and 10\,s windows, respectively. We mean-pool the embedding over the temporal patch dimension $P$, yielding a per-window representation of size $C\times D = 30\times768 = 23{,}040$. 

\subsubsection{Ego-vision features:}
For all video models, the frames of each window were resized along the shorter edge and center-cropped to the native input resolution expected by each model. Frames were sampled at uniformly spaced timestamps across the window, by repeating frames when the window contained fewer than the number required by the model, and encoded jointly, producing a single per-window embedding of size $(N, D)$ aligned with the IMU windows and labels. DINOv3 (ViT-B, $\approx$86\,M parameters) is an image model and was therefore applied to the single center frame of each window at $224\times224$, producing $D=768$. VideoMAE-v2 (base variant, also ViT-B) encodes 16 frames at $224\times224$ into $D=768$, V-JEPA~2 (ViT-L, $\approx$300\,M) encodes 64 frames at $256\times256$ into $D=1024$, padding by repetition at the 2\,s length, and EgoVideo (4-frame variant, InternVideo2 backbone, $\approx$1\,B) encodes 4 frames at $224\times224$ into $D=512$. Frame count is fixed per checkpoint, so temporal coverage, model size and pretraining data covary. DINOv3 and VideoMAE-v2 are the closest controlled pair with the same ViT-B backbone and $D=768$, differing mainly in frame count.

\subsection{Experimental setup}

\begin{figure}[tb]
    \centering
    \includegraphics[width=1\linewidth]{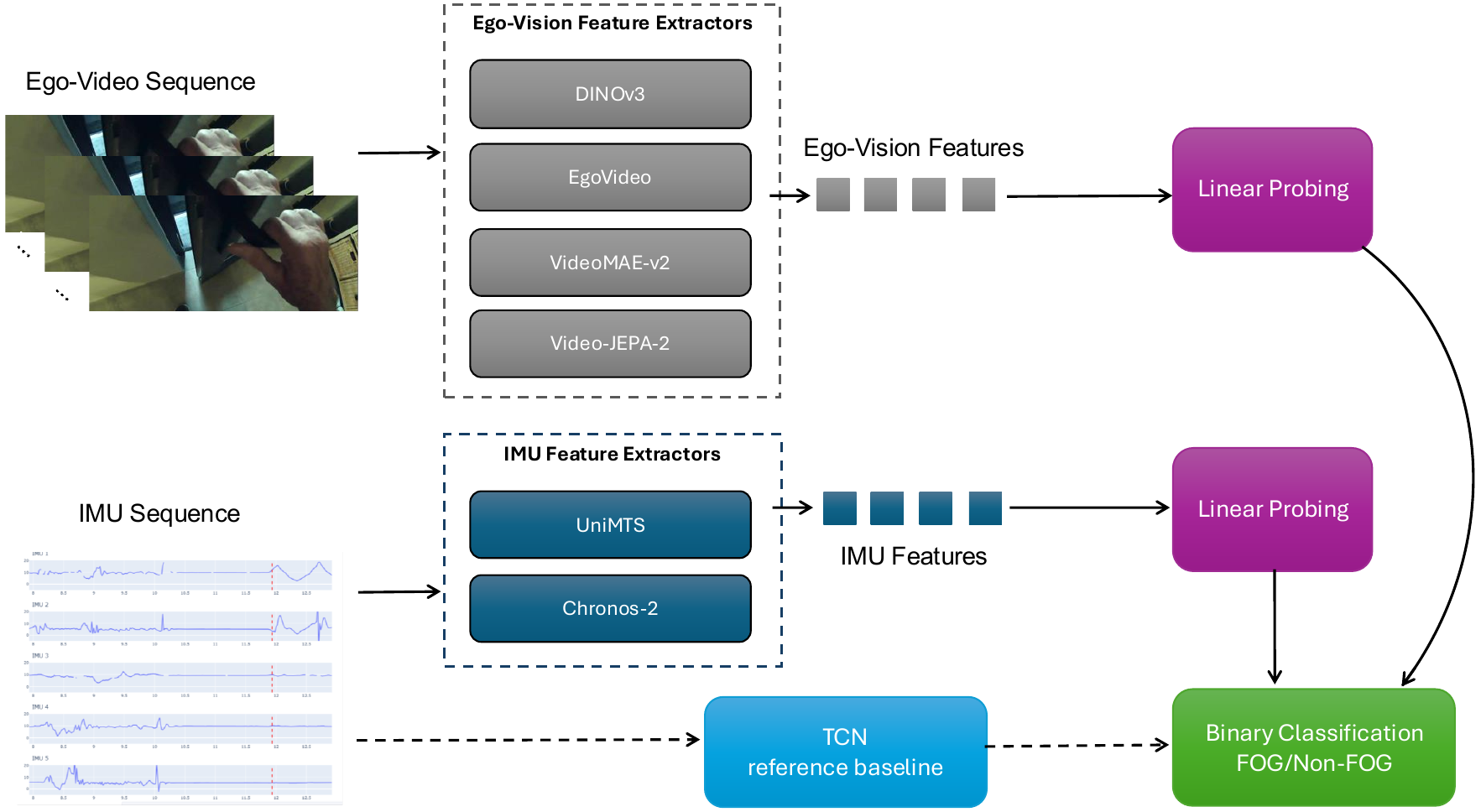}
    \caption{Experimental Setup. For each window, features are extracted from IMU (UniMTS, Chronos-2) and ego-vision (VideoMAE-v2, V-JEPA~2, EgoVideo, DINOv3) foundation models. All models are evaluated under leave-one-subject-out cross-validation, with a fully trained TCN on the IMU signals as a supervised reference.}
    \label{fig:experimental_setup}
\end{figure}

\Cref{fig:experimental_setup} summarizes our experimental pipeline. All models were evaluated under a leave-one-subject-out cross-validation (LOSO-CV) protocol. In each fold, all recordings of one subject, comprising both the ON- and OFF-medication sessions, were held out for testing while the remaining subjects formed the training set. All metrics were computed separately within each held-out subject and then averaged across subjects. The same per-subject averaging was applied to the per-task breakdown over the three ADL tasks (doorway, hotspot, daily life).

For each FM, we assessed its frozen representations with a linear probe, which provides a uniform comparison across models. The per-window embeddings were standardized to zero mean and unit variance using statistics computed on the training folds. An $L_2$-regularized logistic-regression classifier (class-balanced) was then fit on the training windows and used to predict FOG on the held-out subject. We report a single regularization strength $C=10^{-3}$ with all models, chosen from a sweep over $[10^{-6},10]$ in which it remained close to the optimum for every model. Ranking the models at their individual optima instead did not change the order on either F1 or AUROC. We also evaluated the attentive probe used in the original V-JEPA~2 evaluation protocol~\cite{assran2025vjepa2}, which did not improve on the linear probe, likely because our training set is small relative to its capacity; we therefore report linear-probe results throughout.

Finally, as fully trained baselines, we train two TCNs end-to-end on accelerometer signals after gravity removal via a 0.3\,Hz high-pass filter, with one variant additionally including the gyroscope signals. We set the number of dilated residual blocks so the receptive field spans the full window (five, six, and eight blocks for the 2, 3, and 10\,s windows). The remaining hyperparameters were selected by grid search, giving a residual block width of 32, kernel size 3, dropout 0.3, and weight decay $10^{-3}$. The TCN was optimized with a class-weighted cross-entropy loss (AdamW \cite{adamw}, learning rate $5\times10^{-4}$, cosine schedule) under the same LOSO protocol.

\subsection{Evaluation metrics}

FOG detection performance was evaluated using multiple metrics, reflecting the complementary aspects of performance captured by each measure \cite{mancini2025technology}. We use F1-score as the primary metric because it balances precision and recall for the minority FOG class, although it can be lower for subjects with few FOG events. Secondary metrics include AUROC, AUPRC, recall, and false positive rate (FPR). AUROC measures class separability and is relatively insensitive to changes in FOG prevalence, but can appear optimistic in imbalanced settings due to the abundance of non-FOG samples. AUPRC focuses on the FOG class and is therefore more informative under class imbalance, but is sensitive to prediction confidence; even when two models produce the same false positives, the model assigning higher scores to those false positives relative to true positives will achieve lower AUPRC. AUROC and AUPRC were computed from predicted probabilities, while F1-score, recall, and FPR were computed after thresholding predictions at 0.5. Metrics that are undefined for a held-out subject are omitted from the corresponding average rather than scored as zero. Two subjects contribute no FOG-positive window under our labelling rule, so F1, recall, AUPRC and AUROC are averaged over the remaining 11 subjects, while FPR is defined for all 13; the same rule applies within the per-task breakdown. Because the same subjects are excluded for every model, this leaves the model rankings and paired comparisons unchanged.

Statistical analyses were performed on the subject-level scores obtained from LOSO-CV. Differences between modalities or models were first assessed using the Friedman test, a non-parametric repeated-measures alternative to ANOVA. Post-hoc pairwise comparisons were conducted using the Wilcoxon signed-rank test. To control for multiple comparisons $p$-values were adjusted using Holm's correction. Statistical significance was defined as $p < 0.05$.

\section{Results}

\subsection{Ego-vision contains FOG-relevant information}

\Cref{tab:freeliving_window_overall_2s} reports the main comparison at the 2\,s window length across IMU and ego-vision foundation models, alongside the fully trained TCN baselines. Notably, adding gyroscope data to the TCN baseline lowered F1 relative to accelerometer alone (35.5 vs.\ 42.3), but also reduced the number of false positives, as reflected by the reduced FPR (5.8\% vs.\ 7.2\%). Among the IMU FMs, Chronos-2 was the stronger representation (F1 38.7 vs.\ 29.1) and operated at a much lower FPR (6.1\% vs.\ 28.8\%), whereas UniMTS achieved higher recall (56.7\% vs.\ 39.9\%). Ego-video representations were predictive of FOG, most clearly for EgoVideo and V-JEPA2. On F1, the single-frame DINOv3 representation was weakest (17.0), significantly below the trained TCN. Comparing DINOv3 with VideoMAE-v2, which share a backbone and output dimensionality, AUROC rose from 53.6 to 68.6 when moving from one frame to sixteen. All three video encoders exceeded the single-frame baseline on AUROC (68.6, 72.4 and 77.2 vs.\ 53.6; $p<0.01$ in each case), indicating that egocentric video carries FOG-relevant information beyond a single frame. Capacity does not explain these differences: the largest backbone (EgoVideo, $\approx$1\,B) did not outperform V-JEPA~2 ($\approx$300\,M; F1 27.7 vs.\ 32.6). Across all models, the standard deviations are large relative to the mean scores reflecting high between-subject variability.

\begin{table*}[tb]
  \centering
  \caption{Window-level FOG detection on the ADL tasks. Windows are 2 seconds long. Rows in the foundation-model blocks are linear probes on frozen features; the final block reports fully trained TCN IMU baselines. Models are evaluated leave-one-subject-out and metrics reported are mean ($\pm$ std) across held-out subjects. All values are \%. \textbf{Bold}/\underline{underline}: the best and second-best per column for foundation models.}
  \label{tab:freeliving_window_overall_2s}
  \renewcommand{\arraystretch}{1.5}
  \setlength{\tabcolsep}{4pt}
  \resizebox{0.95\textwidth}{!}{
  \begin{tabular}{@{}|c@{\hspace{4pt}}l|c|c|c|c|ccc|c|@{}}
    \hline
    &  &  &  &  &  & \multicolumn{4}{c|}{AUROC$\uparrow$} \\
    \cline{7-10}
    & Model & F1$\uparrow$ & Recall$\uparrow$ & FPR$\downarrow$ & AUPRC$\uparrow$ & D & H & DL & Overall \\
    \hline
    \multicolumn{10}{|l|}{\textit{ Frozen foundation models}} \\
    \hline
    \multirow{2}{*}{\rotatebox[origin=c]{90}{\tiny IMU}}
      & UniMTS  & 29.1 ($\pm$ 15.8) & \textbf{56.7 ($\pm$ 23.8)} & 28.8 ($\pm$ 10.8)* & 28.9 ($\pm$ 16.7)* & 75.0 ($\pm$ 17.2) & 72.8 ($\pm$ 10.6) & 76.9 ($\pm$ 9.2) & 73.5 ($\pm$ 9.4)* \\
      & Chronos-2 & \textbf{38.7 ($\pm$ 17.4)} & 39.9 ($\pm$ 18.8) & \textbf{6.1 ($\pm$ 4.0)} & \textbf{42.0 ($\pm$ 20.7)}* & \textbf{91.1 ($\pm$ 6.4)} & \textbf{81.4 ($\pm$ 10.9)} & \textbf{83.9 ($\pm$ 6.8)} & \textbf{82.9 ($\pm$ 7.5)} \\
    \cline{2-10}
    \multirow{4}{*}{\rotatebox[origin=c]{90}{\tiny Vision}}
      & DINOv3   & 17.0 ($\pm$ 8.9)* & 28.1 ($\pm$ 10.0) & 19.0 ($\pm$ 11.8)* & 15.4 ($\pm$ 10.6)* & 56.4 ($\pm$ 17.6)* & 53.1 ($\pm$ 14.5)* & 58.0 ($\pm$ 10.5) & 53.6 ($\pm$ 7.0)* \\
      & VideoMAE-v2 & 22.0 ($\pm$ 14.1) & 32.8 ($\pm$ 22.6) & 19.3 ($\pm$ 13.1) & 21.8 ($\pm$ 14.0)* & 79.3 ($\pm$ 12.2) & 65.3 ($\pm$ 15.8) & 68.3 ($\pm$ 9.5) & 68.6 ($\pm$ 9.1)* \\
      & EgoVideo & 27.7 ($\pm$ 16.4) & \underline{52.3 ($\pm$ 24.6)} & 26.7 ($\pm$ 17.2)* & 23.9 ($\pm$ 14.2)* & 74.3 ($\pm$ 11.5) & 70.6 ($\pm$ 11.3) & 75.5 ($\pm$ 9.3) & 72.4 ($\pm$ 7.7)* \\
      & V-JEPA2  & \underline{32.6 ($\pm$ 13.6)} & 47.4 ($\pm$ 25.7) & \underline{17.6 ($\pm$ 16.7)} & \underline{33.0 ($\pm$ 16.9)}* & \underline{87.1 ($\pm$ 11.5)} & \underline{77.6 ($\pm$ 10.2)} & \underline{77.1 ($\pm$ 14.2)} & \underline{77.2 ($\pm$ 6.3)}* \\
    \hline
    \multicolumn{10}{|l|}{\textit{Fully trained IMU Baseline}} \\
    \hline
    \multirow{2}{*}{\rotatebox[origin=c]{90}{\tiny IMU}}
      & TCN (acc only)  & 42.3 ($\pm$ 19.8) & 48.6 ($\pm$ 24.6) & 7.2 ($\pm$ 6.2) & 47.3 ($\pm$ 20.7) & 90.1 ($\pm$ 10.1) & 83.5 ($\pm$ 10.5) & 80.6 ($\pm$ 11.1) & 83.0 ($\pm$ 8.8) \\
      & TCN (acc+gyro)  & 35.5 ($\pm$ 21.6) & 36.6 ($\pm$ 25.8) & 5.8 ($\pm$ 4.1) & 40.0 ($\pm$ 23.4) & 87.1 ($\pm$ 17.7) & 78.9 ($\pm$ 14.2) & 80.3 ($\pm$ 11.3) & 83.3 ($\pm$ 7.1) \\
    \hline
    \end{tabular}
    }
  \caption*{\scriptsize Per-task AUROC: D = doorway, H = hotspot, DL = daily life. * indicates a statistically significant difference from TCN (acc only) ($p_{\mathrm{corrected}} < 0.05$, Wilcoxon signed-rank test with Holm correction).}
\end{table*}

Notably, V-JEPA2 was competitive with the IMU FMs. It surpassed the motion-specialized UniMTS on F1 (32.6 vs.\ 29.1), AUROC (77.2 vs.\ 73.5), and FPR (17.6\% vs.\ 28.8\%). Results suggest frozen egocentric video features alone are predictive of FOG at a level comparable to the IMU FMs, with the fully trained TCN (F1 42.3) remaining the strongest reference.

\subsection{Effect of window length}
\Cref{tab:window_length_ablation} examines how the standalone performance of each IMU and ego-video FM changes as the temporal window widens from 2\,s to 10\,s. Across models, F1-score generally increased with window length (\eg, Chronos-2 38.7 to 42.4, V-JEPA2 32.6 to 36.4), likely due to the higher proportion of FOG windows and the resulting reduction in class imbalance (\Cref{fig:total_windows}). In contrast, AUROC consistently decreased (\eg, Chronos-2 82.9 to 71.6, V-JEPA2 77.2 to 66.4), suggesting that longer windows reduce the separability of FOG and non-FOG samples. A likely explanation is that longer windows contain a mixture of FOG and non-FOG behaviors and less precise temporal alignment with FOG events, producing more heterogeneous samples and noisier labels. Consequently, comparisons across window lengths should be interpreted cautiously, as increasing F1 does not necessarily indicate improved discriminative performance. By AUROC, Chronos-2 remained the strongest model at every window length, and V-JEPA2 was the best ego-video representation at the shorter 2 and 3\,s windows.

\begin{table*}[tb]
  \centering
  \caption{Window size ablation for video and IMU foundation models. Values are F1/AUROC (\%) averaged over the LOSO subjects. \textbf{Bold}: best value per column.}
  \label{tab:window_length_ablation}
  \renewcommand{\arraystretch}{1.2}
  \setlength{\tabcolsep}{4pt}
  \scriptsize
  \resizebox{0.8\textwidth}{!}{
  \begin{tabular}{@{}|l|cc|cc|cc|@{}}
    \hline
    & \multicolumn{2}{c|}{2\,s}
    & \multicolumn{2}{c|}{3\,s}
    & \multicolumn{2}{c|}{10\,s} \\
    \cline{2-7}
    Model & F1 & AUROC & F1 & AUROC & F1 & AUROC \\
    \hline
    \multicolumn{7}{|l|}{\textit{IMU foundation models}} \\
    \hline
    UniMTS
      & 29.1 & 73.5
      & 34.4 & 73.4
      & \textbf{45.1} & 67.7 \\
    Chronos-2
      & \textbf{38.7} & \textbf{82.9}
      & \textbf{42.0} & \textbf{82.2}
      & 42.4 & \textbf{71.6} \\
    \hline
    \multicolumn{7}{|l|}{\textit{Ego-video foundation models}} \\
    \hline
    DINOv3
      & 17.0 & 53.6
      & 18.6 & 57.2
      & 31.4 & 54.3 \\
    VideoMAE-v2
      & 22.0 & 68.6
      & 29.9 & 71.4
      & 41.7 & 65.6 \\
    EgoVideo
      & 27.7 & 72.4
      & 32.4 & 71.0
      & 35.5 & 54.4 \\
    V-JEPA2
      & 32.6 & 77.2
      & 34.8 & 74.5
      & 36.4 & 66.4 \\
    \hline
  \end{tabular}
  }
\end{table*}

\subsection{Qualitative Findings}

\begin{figure}[tb]
    \centering
    \includegraphics[width=1\linewidth]{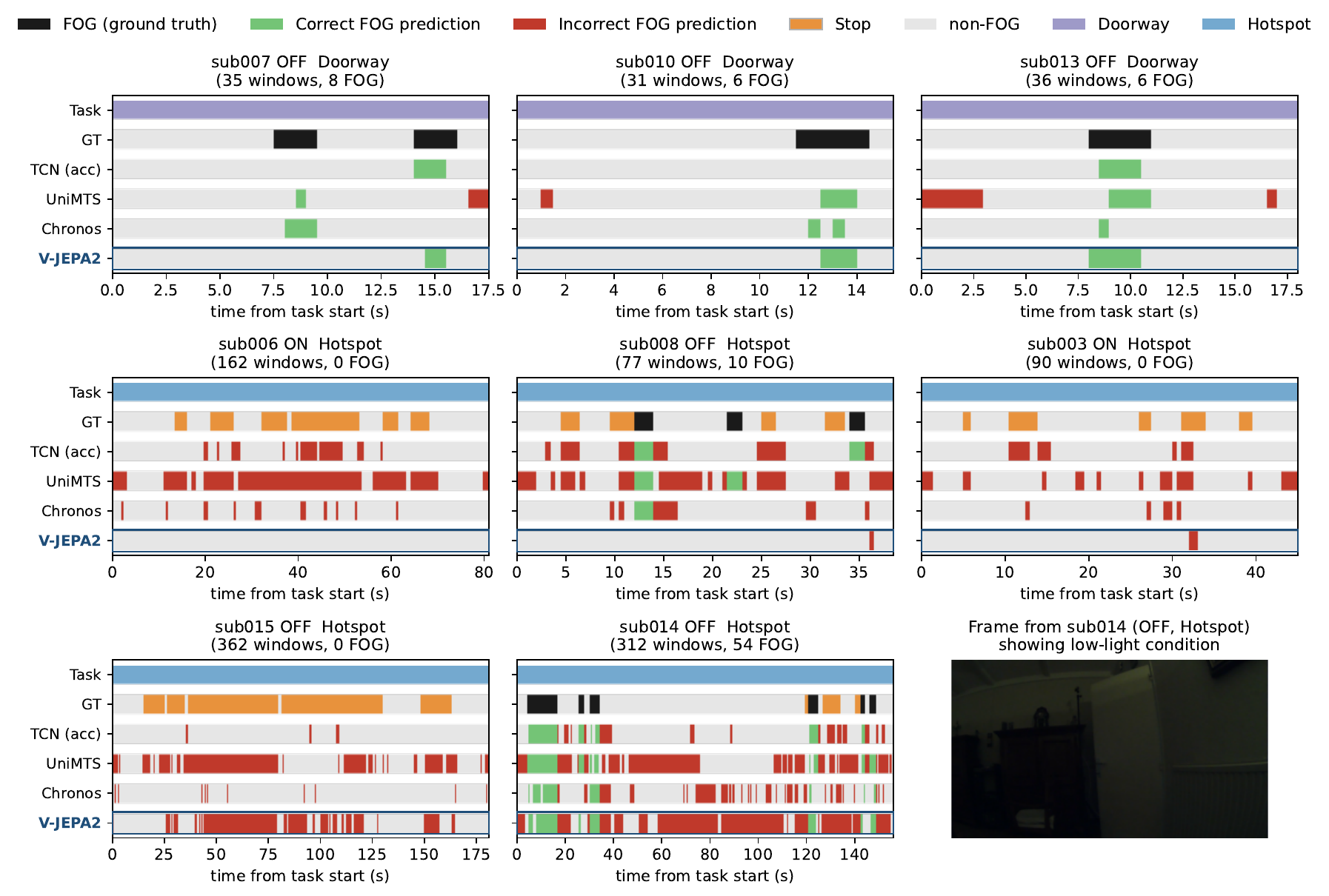}
    \caption{Predictions over time for the ground-truth (GT), the trained TCN, the IMU FMs, and V-JEPA2. Each timestep represents a model prediction made every 0.5\,s using 2\,s windows of input signal. The segments are shown as FOG, Stop, or non-FOG. Stop is treated as non-FOG during training and displayed here for comparison. The ego-vision model mostly agrees with the IMU models (top) and in some cases better separates real freezes from voluntary stops (middle). It does not always help (bottom): false FOG predictions for sub015, and a poorly lit sub014 segment (frame on the right) where the visual features fail. Illustrative examples only.}    
    \label{fig:cases}
\end{figure}

Representative examples of model predictions are shown in \Cref{fig:cases}. In several tasks, both modalities detected FOG episodes at similar time points (\eg, subjects 007, 010 Doorway) suggesting that ego-vision contains independent predictive information. Individual recordings also show segments where the ego-vision model identifies prolonged stopping better than the IMU models (\cref{fig:cases}, middle). This does not hold across the cohort: measuring FPR on windows containing at least 0.5\,s of annotated stopping and no FOG, V-JEPA2 produced more false alarms than both Chronos-2 (24.9\% vs.\ 9.7\%) and the acc-only TCN (11.0\%), and every model false-alarmed more during stopping than on other non-FOG windows (see Supplementary Material). Voluntary stopping is therefore a failure mode shared across modalities, and whether visual context helps appears to vary by recording. The vision model also missed some FOG episodes correctly identified by the IMU models (\eg, subject 008). In a few cases ego-vision performed much worse than the IMU baseline (\eg, subjects 015 and 014). For example, subject 015 had no annotated FOG episodes yet generated many false positives for V-JEPA2, possibly because prolonged stopping occurred during object interaction (\eg, cleaning a cat litter box) while the hands were not visible. Similarly, subject 014 displayed poor performance, potentially due to the low-light recording conditions (\cref{fig:cases}, bottom right), which may have degraded the quality of the extracted visual representations.

Together, these analyses suggest that ego-vision provides information that is not simply redundant with inertial sensing. At the same time, the errors observed under low-light conditions and in subjects with atypical movement patterns show that visual context alone is insufficient, suggesting further research in multimodal models that can combine contextual and kinematic information.

\section{Limitations and Future Work}

This study shows that ego-vision enables context-aware clinical motion understanding during home-based ADL through the use case of FOG detection. We provided evidence that ego-vision features from frozen FM representations carry clinically meaningful motion-context information relevant for FOG detection. However, it did not reduce false positives during voluntary stopping relative to the strongest inertial baselines except in a few subjects. Finally, pretrained FM features can reach similar performance to a TCN trained from scratch, supporting their potential for clinical motion applications where labeled datasets are limited. It remains unclear whether the ego-vision signal reflects motion, environmental context, or both and future work could address this through a deeper interpretability analysis such as applying explainability techniques on the learned feature space.

These findings should be interpreted in light of the scale of the dataset. Collecting synchronized multimodal recordings from people with PD at homes is technically and clinically challenging, particularly when recordings include both medication states, multiple wearable sensors, ego-video, and expert video-based FOG annotation. As a result, the present dataset includes a relatively small cohort and short semi-structured ADLs. Although this setting captures more real-world behavior than laboratory protocols and includes substantial variability in FOG severity and home environments, longer unsupervised free-living recordings are needed to evaluate robustness across broader daily-life conditions. Furthermore, while no participants reported concerns regarding the smart glasses, their long-term usability and acceptability require further investigation, particularly given the potential practical and privacy considerations in everyday use.

Beyond data scale, our framework relies solely on frozen features from pretrained FMs. While this is efficient in a small-data regime and avoids overfitting, it leaves the encoders unadapted to FOG, and fine-tuning strategies such as low-rank adaptation (LoRA) \cite{hu2022lora} may yield further gains. The results suggest complementary error patterns across modalities, though this remains preliminary since fusion of IMU and ego-vision was not explored here, motivating future work on multimodal fusion and prediction-refinement approaches \cite{xu2023multimodal,yang2025multimodal}. Finally, the current pipeline performs offline classification over pre-segmented windows and is not directly applicable to real-time use, whereas practical home monitoring and closed-loop interventions ultimately require online, low-latency inference. Bridging this gap will require adapting the approach to streaming, causal prediction and developing lightweight or distilled encoders that can run continuously on wearable or on-device hardware \cite{qu2025mobile}. 

\section{Conclusion} 
This study presents the first use of ego-vision for FOG detection and demonstrates its potential for context-aware motion understanding during home-based ADLs. Using frozen representations from pretrained video and IMU foundation models with a simple linear probe, we achieved performance comparable to a fully trained TCN baseline, suggesting that clinically relevant predictive information is already present in these representations. Furthermore, ego-vision alone performed competitively with IMU-based models, providing evidence that visual context contains independent cues relevant to FOG. A direct test on annotated stop windows, however, showed that ego-vision FM features alone do not disambiguate voluntary stopping from freezing better than inertial sensing. Together, these findings motivate future work on context-aware multimodal approaches, including FM fine-tuning, fusion strategies, and real-time deployment for home-based clinical motion monitoring.

\section*{Acknowledgements}
We thank all participants who gave their time and effort to participate in the study. This study was funded, in part, by the AidWear project funded by the Federal Public Service for Policy and Support, the AID-FOG project by the Michael J. Fox Foundation for Parkinson’s Research under Grant No.: MJFF-024628, the strategic basic research project RevalExo (S001024N) funded by the Research Foundation Flanders, and the Flemish Government under the Flanders AI Research Program (FAIR). The resources and services used in this work were provided by the VSC (Flemish Supercomputer Center), funded by the Research Foundation - Flanders (FWO) and the Flemish Government.

%
%
\bibliographystyle{splncs04}
\bibliography{main}
\end{document}


\titlerunning{Supplementary material}

\title{Supplementary Material for "Towards Context-Aware Clinical Motion Understanding in Daily Living at Home: Freezing of Gait Detection with Egocentric Vision"}
\author{Vayalet Stefanova\inst{1}\orcidlink{0009-0006-0137-5033}$^{\star}$ \and
Diwas Lamsal\inst{1}\orcidlink{0009-0006-1125-2299}$^{\star}$ \and
Margot Genbrugge\inst{2}\orcidlink{0009-0002-8380-7290} \and
Maxim Yudayev\inst{1}\orcidlink{0000-0001-9521-1537} \and
Christian Schlenstedt\inst{3}\orcidlink{0000-0002-3838-6848} \and 
Moran Gilat\inst{2}\orcidlink{0000-0002-0072-4637} \and
Bart Vanrumste\inst{1}\orcidlink{0000-0002-9409-935X}$^{\dagger}$ \and
Benjamin Filtjens\inst{4, 5}\orcidlink{0000-0003-2609-6883}$^{\dagger}$}
\authorrunning{V.~Stefanova, D.~Lamsal et al.}

\institute{Department of Electrical Engineering (ESAT), KU Leuven, Leuven, Belgium \and
Neurorehabilitation Research Group (eNRGy), Department of Rehabilitation Sciences, KU Leuven, Leuven, Belgium \and 
Institute of Interdisciplinary Exercise Science and Sports Medicine, Medical School Hamburg, Hamburg, Germany \and
Department of Engineering Systems and Services, Delft University of Technology, Delft, The Netherlands \and
Institute for Health Systems Science, Delft University of Technology, Delft, The Netherlands}
\maketitle

\section{Complete list of daily life tasks}
The data collection protocol for each participant included five activities of daily living: walking through a doorway, two daily-life tasks, and two hotspot tasks. The two daily-life tasks were randomly selected from the following list of 14 predefined tasks:
\begin{enumerate}
    \item Set the dining table for two people, sit at the table, then clear the table.
    \item Walk from the couch/chair to the kitchen, fill a glass with water, get a snack, and return to the couch/chair.
    \item (Fictively) water the plants in the living room.
    \item Walk to every corner of the living room.
    \item Turn on the main lights in all rooms tested.
    \item Walk around the coffee table/dining table in both directions.
    \item Walk to the front door, open and close it, then return to the living room.
    \item Walk from the couch/chair to the table, get a book/newspaper/tablet, walk back to the couch/chair, and sit down.
    \item Empty and fill the dishwasher:
    \begin{enumerate}
        \item If no dishwasher is present: (fictively) wash some dishes and put the clean tableware in its correct location
    \end{enumerate}
    \item Close and open the curtains for all windows in all rooms.
    \item Tidy up the rooms.
    \item Go look at two art pieces or pictures in your house.
    \item Clean/dust off some surfaces or furniture in the rooms.
    \item Get a pair of shoes, put them next to the couch, sit on the couch, get up, and put the shoes back.
\end{enumerate}

\section{Effect of window stride}
Our main configuration uses a 2\,s window with a 0.5\,s stride, meaning consecutive windows overlap by 75\%. Combined with the 0.5\,s minimum-overlap FOG criterion and a median FOG episode duration of 0.9\,s, a single FOG episode can generate multiple overlapping windows with varying degrees of temporal purity (i.e., varying proportions of FOG vs.\ non-FOG content within the window). To assess whether this affects reported performance, we evaluated all models at strides of 0.5\,s, 1\,s, and 1.5\,s, corresponding to 75\%, 50\%, and 25\% window overlap, respectively, while keeping all other hyperparameters fixed at the values optimized for the 0.5\,s stride. The resulting number of windows per stride are displayed in \Cref{tab:window_counts} which shows that the number of FOG windows decreases roughly in proportion with the overall window count as stride increases.

\Cref{tab:stride_sensitivity} shows that performance remains broadly stable across strides for most models, with F1 and AUROC typically varying by only a few points as overlap decreases from 75\% to 25\%. 


\begin{table}[tb]
  \centering
  \caption{Window counts across strides.}
  \label{tab:window_counts}
  \begin{tabular}{@{}|l|c|c|c|@{}}
    \hline
    \textbf{Stride} & \textbf{0.5\,s} & \textbf{1.0\,s} & \textbf{1.5\,s} \\
    \hline
    Overlap     & 75\%     & 50\%    & 25\%    \\
    Windows     & 10{,}249 & 5{,}118 & 3{,}420 \\
    FOG windows & 1{,}190  & 605     & 410     \\
    \hline
  \end{tabular}
\end{table}

\begin{table*}[tb]
  \centering
  \caption{Window overlap ablation for all models (0.5\,s, 1\,s, 1.5\,s strides; 75\%, 50\%, 25\% overlap). Values are F1/AUROC (\%) averaged over the LOSO subjects. \textbf{Bold}: best value per column; \underline{underline}: second-best.}
  \label{tab:stride_sensitivity}
  \renewcommand{\arraystretch}{1.2}
  \setlength{\tabcolsep}{4pt}
  \scriptsize
  \resizebox{0.8\textwidth}{!}{
  \begin{tabular}{@{}|l|cc|cc|cc|@{}}
    \hline
    & \multicolumn{2}{c|}{0.5\,s (75\%)}
    & \multicolumn{2}{c|}{1.0\,s (50\%)}
    & \multicolumn{2}{c|}{1.5\,s (25\%)} \\
    \cline{2-7}
    Model & F1 & AUROC & F1 & AUROC & F1 & AUROC \\
    \hline
    \multicolumn{7}{|l|}{\textit{IMU foundation models}} \\
    \hline
    UniMTS
      & 29.1 & 73.5
      & 29.1 & 73.0
      & 27.4 & 71.3 \\
    Chronos-2
      & \underline{38.7} & 82.9
      & 36.5 & 82.2
      & 36.6 & 82.2 \\
    \hline
    \multicolumn{7}{|l|}{\textit{Ego-video foundation models}} \\
    \hline
    DINOv3
      & 17.0 & 53.6
      & 14.9 & 52.2
      & 15.6 & 54.3 \\
    VideoMAE-v2
      & 22.0 & 68.6
      & 22.6 & 68.5
      & 23.5 & 68.6 \\
    EgoVideo
      & 27.7 & 72.4
      & 27.9 & 71.3
      & 29.0 & 72.2 \\
    V-JEPA2
      & 32.6 & 77.2
      & 31.6 & 77.4
      & 31.5 & 78.0 \\
    \hline
    \multicolumn{7}{|l|}{\textit{Fully trained IMU Baseline}} \\
    \hline
    TCN (acc only)
      & \textbf{42.3} & \underline{83.0}
      & \underline{38.4} & \underline{82.6}
      & \textbf{38.7} & \textbf{82.8} \\
    TCN (acc+gyro)
      & 35.5 & \textbf{83.3}
      & \textbf{40.3} & \textbf{84.4}
      & \underline{38.4} & \underline{82.3} \\
    \hline
  \end{tabular}
  }
\end{table*}

\section{False alarms during voluntary stopping}
A stop window is defined as one containing at least 0.5\,s of annotated voluntary stopping and no FOG, mirroring the FOG criterion, giving 2{,}024 windows across all 13 subjects. \Cref{tab:stopfpr} reports FPR restricted to these windows and to the remaining non-FOG windows, computed per subject and averaged and the ordering follows each model's overall false-alarm rate. Every model false-alarmed more on stop windows than on other non-FOG windows showing the difficulty in differentiating FOG episodes from voluntary stops. V-JEPA2 (24.9\%) did not improve on Chronos-2 (9.7\%) or the trained TCN (11.0\%), which may partly reflect the ego-video models' generally lower overall performance relative to the IMU-based baselines. Whether combining visual and inertial modalities could yield further improvement remains an open question for future work.

\begin{table}[tb]
  \centering
  \caption{False positive rate on annotated voluntary-stop windows compared with other
  non-FOG windows, at the 2\,s window length. Values are \%, computed per subject and
  averaged over the 13 LOSO subjects. * indicates a significant difference from TCN
  (acc only) ($p_{\mathrm{corrected}}<0.05$, Wilcoxon signed-rank test with Holm
  correction).}
  \label{tab:stopfpr}
  \renewcommand{\arraystretch}{1.2}
  \begin{tabular}{@{}|l|c|c|@{}}
    \hline
    \textbf{Model} & \textbf{FPR stop}$\downarrow$ & \textbf{FPR non-stop}$\downarrow$ \\
    \hline
    TCN (acc+gyro)  &  7.1  &  4.8 \\
    Chronos-2       &  9.7  &  4.4 \\
    TCN (acc only)  & 11.0  &  5.4 \\
    DINOv3          & 22.3* & 17.6 \\
    V-JEPA2         & 24.9  & 13.8 \\
    VideoMAE-v2     & 33.3* & 15.0 \\
    EgoVideo        & 44.5* & 20.5 \\
    UniMTS          & 49.1* & 21.7 \\
    \hline
  \end{tabular}
\end{table}